\documentclass[a4paper, 10pt, conference]{ieeeconf}      

\IEEEoverridecommandlockouts                              

\usepackage{amsmath} 
\usepackage{graphicx}
\usepackage{algorithm}
\usepackage{algpseudocode}
\usepackage{makecell}
\usepackage{listings}
\usepackage{xcolor}

\usepackage[unicode]{hyperref}
\usepackage{hyperxmp}
\hypersetup{%
  pdftitle={UNCOM: Zero-shot Context-Aware Command Understanding for Tabletop Scenarios},
  pdfauthor={Antonio Galiza Cerdeira Gonzalez; Pawe\l{} Gajewski; Bipin Indurkhya},
  pdfsubject={Robot command understanding; zero-shot learning; context-aware language grounding},
  pdfkeywords={UNCOM, zero-shot, context-aware, command understanding, tabletop, robotics, language grounding},
  pdfcreator={LaTeX},
  pdfproducer={pdfTeX},
  pdfdisplaydoctitle=true,
  colorlinks=true,
  linkcolor=blue,
  citecolor=blue,
  urlcolor=blue
}

\lstdefinestyle{pythonstyle}{
    language=Python,
    basicstyle=\ttfamily\footnotesize,
    keywordstyle=\bfseries\color{blue},
    commentstyle=\itshape\color{green!50!black},
    stringstyle=\color{orange},
    numbers=left,
    numberstyle=\tiny,
    stepnumber=1,
    numbersep=5pt,
    frame=single,
    breaklines=true,
    showspaces=false,
    showstringspaces=false
}

\title{\LARGE \bf
UNCOM: Zero-shot Context-Aware Command Understanding for Tabletop Scenarios
}

\author{Antonio Galiza Cerdeira Gonzalez$^{1, 2}$, Paweł Gajewski$^{2,3}$ and Bipin Indurkhya$^{2}$
\thanks{$^{1}$Corresponding author, {\tt\small antonio.gonzalez@uj.edu.pl}}%
\thanks{$^{2}$Author is with Jagiellonian University, Krakow, Poland}%
\thanks{$^{3}$Author is with AGH University of Krakow, Poland}%
}

\begin{document}

\maketitle
\thispagestyle{empty}
\pagestyle{empty}

\begin{abstract}

This paper presents UNCOM, a novel hybrid framework for interpreting natural human commands in tabletop scenarios. The system integrates multiple sources of information -- speech, gestures, and scene context -- to extract structured, actionable instructions for robots. Addressing the need for general-purpose human-robot interaction in domestic environments, UNCOM is designed for zero-shot operation, without reliance on predefined object models or training data specific to a given task. Using foundational and task-specific deep learning models, it allows out-of-the-box speech recognition, natural language understanding, gesture detection, and object segmentation. The modular architecture enhances transparency and explainability by explicitly parsing commands into object-action-target representations, enabling integration with symbolic robotic frameworks. We demonstrate the system in a TIAGo++ robot and provide an evaluation on a real-world data set of human-robot interaction scenarios; achieving an 82.39\% success rate over our benchmark data set, highlighting the robustness of the system to diversity, noise, and communication ambiguity. The data set, evaluation scenarios, and the code are publicly available to support future research.

\end{abstract}


\section{Introduction}

Current household service robots mainly perform one task, such as vacuuming, lawn-mowing, or window-cleaning. If these robots are to become more capable, seamless communication with users of varying degrees of expertise is essential. Human natural language has both verbal and gestural components, and understanding the latter is critical to reducing the ambiguity of instruction. 

Since human natural communication is rich in verbal and body cues, multimodal command understanding remains a central problem in human-robot interaction (HRI) research.  The complexity stems from the need to correctly interpret natural speech and visual cues, which can be ambiguous, diverse, and include noise in real-world settings. People also have different communication styles and point in distinctive ways~\cite{tymon_gestures}. Furthermore, traditional object recognition systems are trained on datasets that contain a subset of the myriad of objects and contexts offered by reality, which leads to several bottlenecks in system performance. 
To address these challenges, we have developed UNCOM  (\textbf{UN}derstanding \textbf{COM}mands) -- a hybrid system architecture that leverages foundational models and a smaller specialist model for transcribing speech and for gesture detection in tabletop scenarios. The deployment and flow of data between the models is performed in Python with a simple algorithm focused on modularity. The larger foundational models (Phi-4, Whisper-large-v3-turbo, GroundingDINO, DINOv2) are responsible for complex sub-tasks such as automatic speech recognition, natural language understanding, and general object detection. The MediaPipe Hand Landmark model is used for gesture detection, being the only specialist model in the core architecture. Our system, then, operates in a fully zero-shot manner, requiring no fine-tuning with specialized household service datasets. This decoupling between understanding/reasoning and execution and the zero-shot capabilities allows our system to effectively function across a wide range of environments, tasks, and robot embodiments, requiring UNCOM adopters to develop task execution sub-systems, which might be specific to their robot embodiment. 

\begin{figure*}[thpb]
  \centering
  \includegraphics[width=0.45\linewidth]{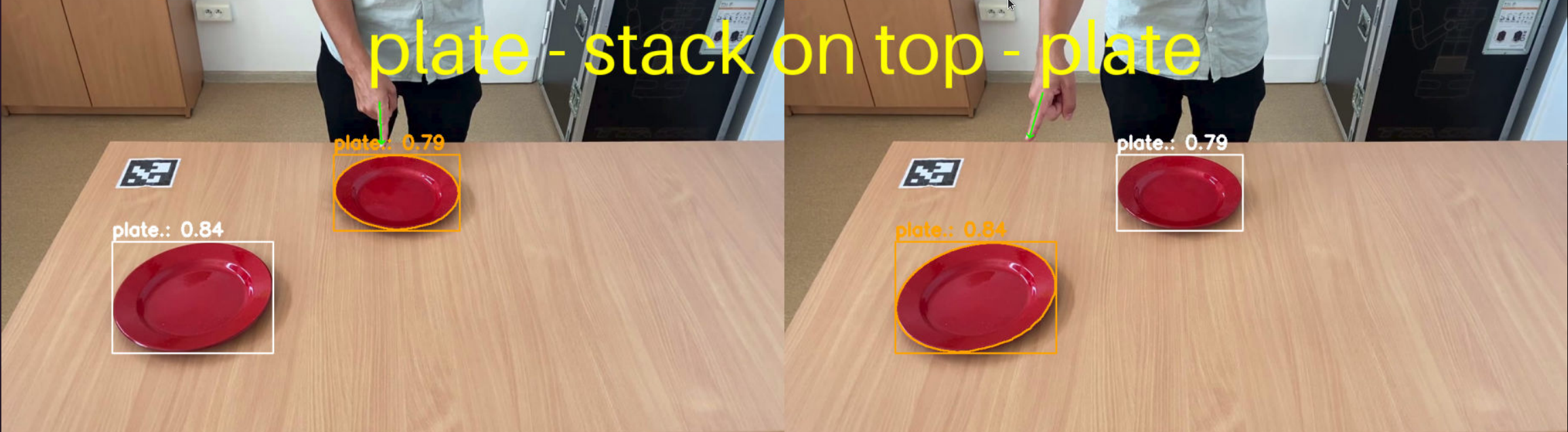}~\includegraphics[width=0.45\linewidth]{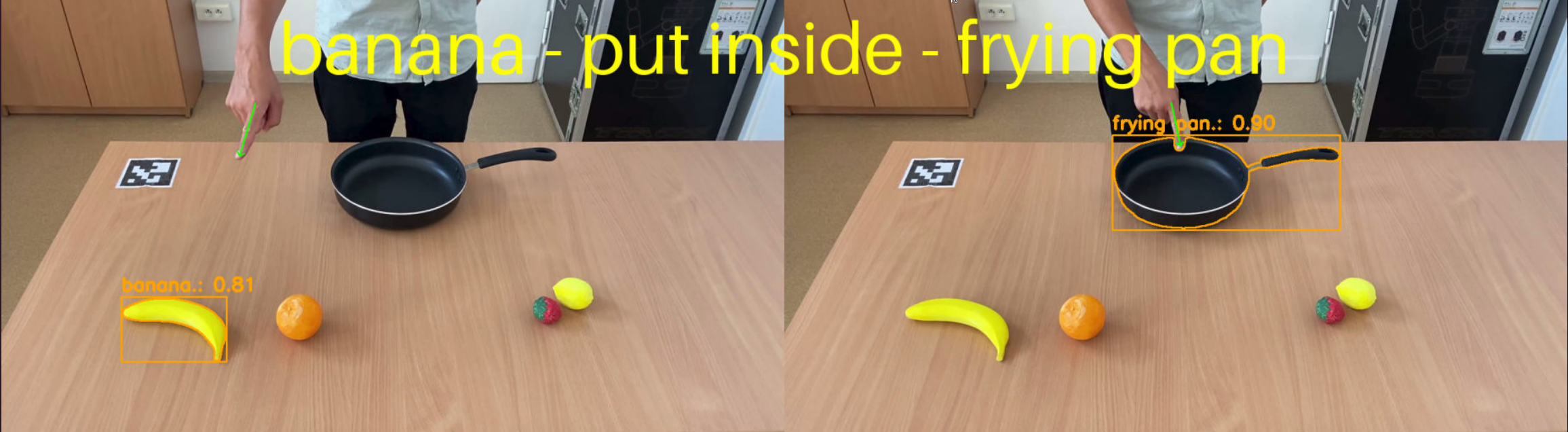}
  \caption{Automatically generated command annotations created from videos. Text indicates extracted elements of the command in the format: object - action - target. Detected objects that are relevant to the command are highlighted on the table, with the selected object highlighted in orange and segmented out, with a contour marked around it. If pointing is detected, the pointing vector is highlighted in green. In the top example, the command was: “Take this plate and stack it on top of the other plate.” In the bottom, the command was: “Take the banana and put it inside the frying pan.”}
  \label{fig:examples}
  \vspace{-.5 cm}
\end{figure*}

In addition to zero shot operation, UNCOM provides distilled and actionable information to other planning modules (see Figure \ref{fig:examples}) in a larger robotic system by parsing natural language instructions into structured representations that explicitly define target objects and the corresponding actions. It also enhances the explainability of the system's actions, unlike larger monolithic VLA models that transform commands and visual information directly into robot action. UNCOM also assists in other tasks, such as tracking, manipulating, and grasping, by segmenting objects of interest. In the experimental evaluation of our work,  our system achieved an 82.39\% success rate (131 of 159 instances) in our data set. For robot demonstrations, a GDRNet++ model fine-tuned for YCB-V objects 6DoF pose estimation was used to provide more precise poses and grasps for task execution, even though our system is capable of grasping \textit{a priori} unknown objects by obtaining their centroids in the 3D space, but it fails for more complex shapes. 


The main contributions of our work include:
\begin{itemize}
\item Architecture of the UNCOM system, a zero-shot hybrid framework for command understanding in tabletop scenarios that is capable of interpreting multimodal (speech, gestures and scene context) commands by leveraging foundational models and MediaPipe Hand Landmarker to generate interpretable distilled and actionable commands;

\item Demonstration of the efficacy of our system in the TIAGo and TIAGo++ robot platforms, as well as a more detailed benchmark of the platform using a small data set, over which our system achieved an efficiency rate of $82.39\%$; 

\item Benchmark data set to evaluate reasoner performance, containing 159 videos of 22 tabletop test scenarios.

\end{itemize}

\section{Related Work}

Household service robots capable of interacting with non-expert users require bridging the gap between multimodal natural human communication and robotic execution. This challenge is addressed mostly through two competing paradigms: monolithic end-to-end models and modular symbolic frameworks.

\paragraph{Monolithic vs. Symbolic Frameworks}
Recent trends in robotics have led to large-scale end-to-end learning models, such as Vision-Language-Action (VLA) architectures \cite{openvla}. These models offer a direct mapping from multimodal input to robot control and can acquire manipulation skills from demonstration data, but they are often characterized by high computational overhead and "black-box" decision-making. In contrast, symbol-based robotic frameworks \cite{adapting, graph_task, sympolic_survey} rely on modular subsystems that parse commands into structured representations. This approach provides inherent transparency and explainability \cite{explainable1, explainable2}, which help establish user trust and safety in domestic environments. UNCOM adopts this modular philosophy, prioritizing intermediate interpretability over monolithic inference, at the cost of requiring per-embodiment execution primitives and a less flexible task repertoire. 

\paragraph{Multimodal Command Understanding}
Human communication relies on speech and gestural cues to resolve ambiguity. Early research on voice and gesture-driven interfaces \cite{old} established the foundation for this field. In the context of navigation, systems like that proposed by Kumar et al. \cite{kumar2021sharing} utilize whole-body analysis to ground robot movement commands. However, despite the zero-shot capacity of the language understanding module used, their system relied on an \textit{a priori} known 2D-occupancy map of the environment, which limits its use by others, unlike our system, which uses DINOv2 (with depth head) and Voronoi diagrams to partition any table and automatically generate occupancy maps.

Several tabletop interfaces have attempted to resolve user intent through specialized GUIs \cite{push} or Bayesian filtering for disambiguation \cite{whitney2016interpreting}. Matuszek et al. \cite{matuszek2014learning} used short-range pointing for intent clarification; such systems often require users to adopt specific pointing styles to compensate for sensor inaccuracies \cite{mayer2018effect, ahn2018interactive}. More recent efforts \cite{weerakoon_gesture_2020} have used datasets like \cite{scalise_natural_2018} to interpret unclear instructions, but these remain largely restricted to simplified environments with limited object sets (e.g., colored blocks). UNCOM uses the gesture for disambiguation, but by leveraging GroundingDINO and MediaPipe's Hand Landmarker, several pointing styles are supported.

\paragraph{Leveraging Foundational Models}

The primary bottleneck for traditional symbolic systems has been the reliance on closed-vocabulary object detectors and hand-coded semantic graphs \cite{forklift}. Although some systems attempt to learn concepts from scratch \cite{learning}, they struggle with the diversity of real-world household objects. UNCOM addresses these limitations by integrating modern foundational models. Using a lightweight reasoner (Phi-4) and an open-vocabulary vision backbone (GroundingDINO), our system enables zero-shot, out-of-the-box operation across diverse tabletop tasks, providing a level of generalization that previous specialized models could not achieve.

\section{Architecture of UNCOM}

\begin{figure}[thpb]
  \centering
  \vspace{-.25cm}
  \includegraphics[width=.5\linewidth]{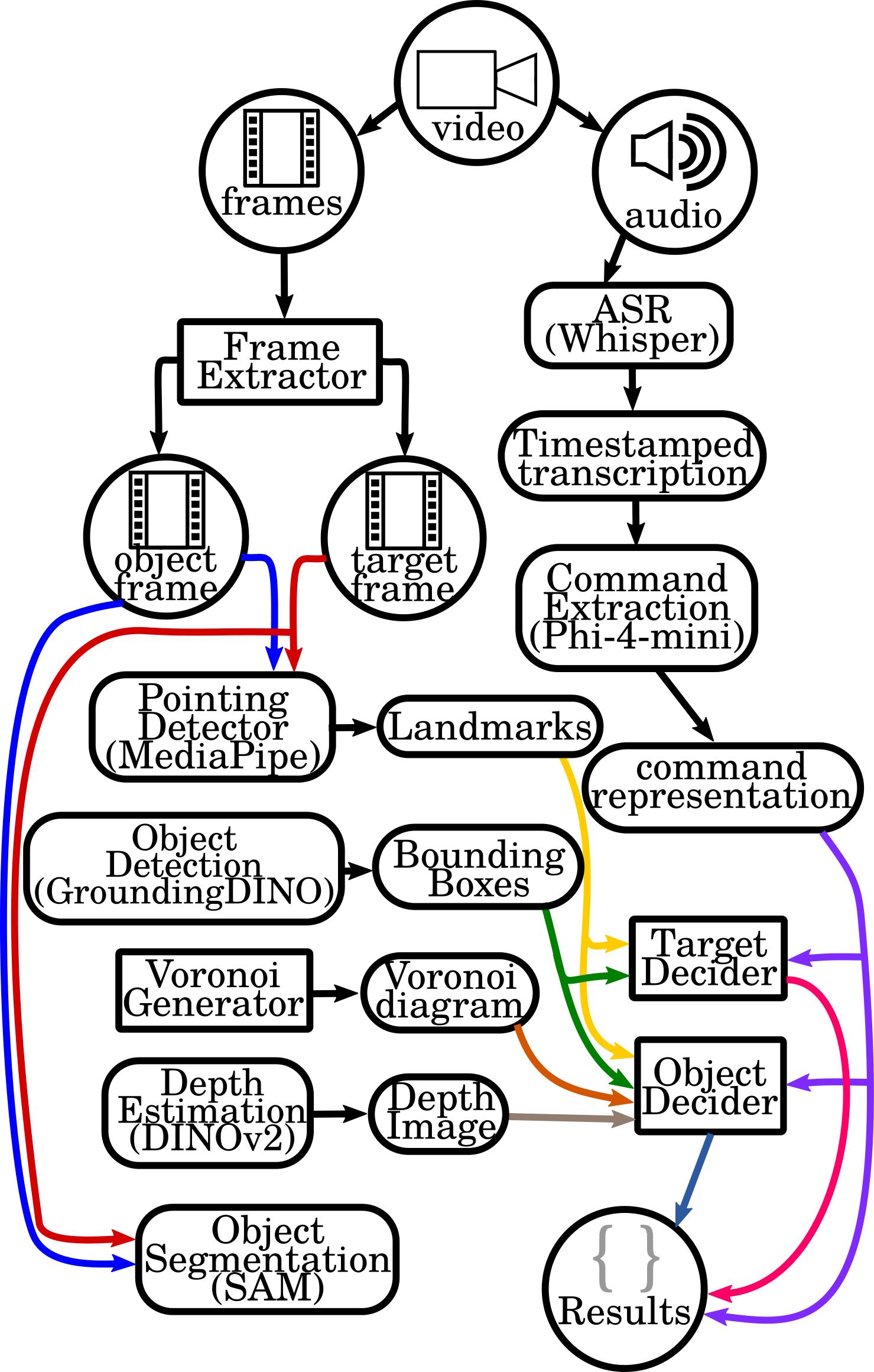}
  \caption{Overview of the algorithm's information flow and general architecture. Square-cornered boxes represent procedural decision-making components, while rounded boxes indicate deep learning models. Arrows were colored to make the diagram easier to follow. The algorithm's pseudocode is available in our public repository (see footnotes).}
  \vspace{-.5cm}
  \label{architecture}
\end{figure}

Our algorithm adopts a structured decision-making framework that employs deep learning models to extract essential information from input data. Audio Transcription is performed by Whisper (large-v3-turbo), command extraction is performed by Phi-4 (mini-instruct), Pointing vector detection is performed by MediaPipe's Hand Landmarker, GroundingDINO (tiny) is used for object detection, object segmentation is achieved by Segment Anything (vit-base), and depth estimation is performed through DINOv2. Figure~\ref{architecture} illustrates a visual representation of the information flow between the core components of our system, whose source code is publicly available\footnote{https://github.com/ichores-research/uncom}. The subsequent sections provide a detailed account of the complete process.

\subsection{Input Processing}

The initial input consists of a video of a tabletop scenario in which users give verbal commands and possibly use pointing gestures. The audio is extracted from the video for further processing.

\subsection{Audio Transcription}

The audio is transcribed using the Whisper-large v3-turbo model\cite{whisper}, configured for English automatic speech recognition (ASR), but other languages or models can be used.  

\subsection{Textual Command Understanding}

The resulting time-stamped JSON transcription is analyzed using the Phi-4 mini-instruct model \cite{phi-4}, which extracts the core semantic elements of the command. That is, the object of interest (noun and adjectives), the intended action (a verb or verb phrase), and the target of that action (noun and adjectives). It also determines whether the object and target are explicitly mentioned or indicated via deixis (e.g., `here', `this').

For that purpose, two sequential prompts are used. The first prompt instructs the model to identify and extract the object, action, and target from the Whisper-generated transcription. The second prompt refines the output by adding details on whether the object and target refer to tangible entities.
The first prompt:

\begin{quote}
You will receive a Whisper JSON transcription. Extract exactly one clear instance of:
- an object (noun + optional adjectives) or a reference to one, like `this', `that', `it';
- an action (verb or phrasal verb),
- a target (noun + optional adjectives, including any positional descriptors like `next to', `between', `near', etc.), or a reference to a target, like `this', `here' or `there'.

Return a single JSON object with the keys: `object', `action', and `target'. Each key should contain a `text' string and a single `timestamp' [start, finish] from the original transcription. If an element is missing or unclear, leave its value empty.

Follow this example strictly:

{`object': {`text': `red mug', `timestamp': [1.04, 1.36]}, `action': {`text': `put on top', `timestamp': [1.5, 1.76]}, `target': {`text': `laptop', `timestamp': [2.24, 2.46]}}

Output only the valid formatted JSON structure. No extra text or formatting.
\end{quote}

The second prompt:

\begin{quote}
Refine your own output to include information whether the object and the target are concrete, specific entities (e.g., `apple', `chair') or a non-concrete reference such as `this', `that', `it'.

Add a field `concrete': true if the object/target is a specific, tangible noun, and `concrete': false if it is a vague or referential term like `this'.

Ensure that pronouns without a clearly identifiable object are marked as `concrete': false.
\end{quote}

\subsection{Video Processing}

After identifying the object and target, the corresponding video frames are extracted using the timestamps obtained during the Speech recognition stage. For the object and the target, a frame corresponding to the timestamp in which the user finished articulating is analyzed for user pointing gestures and object/location detection.  

\subsection{Pointing Gesture Interpretation}

Hand detection is performed on both extracted frames using the MediaPipe Hand Landmarker \cite{mediapipe}. The pointing vector is defined as the line that extends from the base to the tip of the index finger (landmarks 5 to 8). The detector is configured to recognize a maximum of two hands. Additionally, the MediaPipe model estimates the z-coordinate, representing the distance of the hand from the camera. To determine which hand is executing the pointing gesture, the z-coordinate of the index finger is used under the assumption that the hand closer to the camera is the one performing the gesture. While this constitutes a limitation, the system can be expanded to other pointing styles as needed by considering other landmarks, but this is left as future work. 

Whenever the Hand Landmarker fails to detect a pointing vector in the object or in the target frame, the algorithm extracts three previous and next frames in increments of $0.01s$. If no gesture is detected still, the algorithm considers that no pointing gesture could be detected. 

\subsection{Object Detection}

If the object is identified by its name, the name is used as a trigger for the GroundingDINO object detector \cite{grounding_dino}. If GroundingDINO detects multiple objects (or the user refers to the object through a deixis), the algorithm selects the one whose bounding box lies closer to the trajectory defined by the pointing vector.

\subsection{Target Detection}

If the target is referred to by its name, a process similar to object detection is used. 

In contrast, if the target implies a spatial position relative to another object (\textit{e.g.} to the left of this apple), it is usually an unoccupied region on the table. The algorithm proceeds as follows. i) the table is segmented using the object detector, ii) a Voronoi diagram is applied to partition the table into cells, and iii) DINOv2 detects the general `objects' prompt to identify existing objects on the table. The centers of the detected objects, along with the determined reference point, are used to mark overlapping Voronoi cells as occupied, creating an occupancy map of the table. An unoccupied cell in the direction implied by the spatial term is then selected as the target location. This approach allows for tables of any shape, and for cells of different sizes, unlike a standard fixed size grid.

For deictic references, specified solely through pointing, the algorithm first attempts to detect a `container' in the frame. If found, this container is used as the target. When multiple containers are identified, the one closest to the pointing vector is selected. If no container is detected, the target is assumed to be the empty Voronoi cell that is the closest to the pointing gesture.

\subsection{Compiling the Command Representation}

At this stage, the precise pixel regions corresponding to the object and the target are segmented using the Segment Anything Model (SAM) \cite{sam}. The segmentation is guided by the center points of the respective bounding boxes.

The output of this stage consists of the segmented image regions along with the extracted reference nouns for the object, action, and target, forming the complete command representation produced by the algorithm.

\subsection{Applying the Command Representation}

The algorithm yields pixel-level segmentations of both the object and the target from earlier video frames, which can be leveraged by a robot to identify and track these elements within real-time camera feeds. This facilitates accurate localization and interaction with the relevant entities in the current scene.

Then, the extracted commands can be sent to the task execution subsystems, which are beyond the scope of the present architecture, but we use MoveIt! to execute simple pick and place tasks for system evaluation.  

\section{Evaluation of UNCOM}

With the inner workings of UNCOM explained in previous sections, it is necessary to benchmark its performance with data that closely mimic real world applications. 

To evaluate UNCOM's capabilities, we performed two batches of experiments: 

\begin{enumerate}
    \item Pick and place tests on TIAGo and TIAGo ++ robots; 
    \item Comprehensive benchmarking with a dataset of tabletop natural command videos. 
\end{enumerate}

These are explained, respectively, in Sections~\ref{subsec:robot} and  \ref{subsec:benchmarking}. 

\subsection{Demonstration on Robots}~\label{subsec:robot}

To demonstrate that UNCOM can be effectively deployed in tabletop robotics applications, we developed an integrated data acquisition and control interface for the PAL Robotics TIAGo and TIAGo++ platforms, covering camera, microphone, and actuation streams. The current prototype targets TIAGo-class robots and focuses on pick-and-place as a representative task class, though UNCOM itself understands a broader range of candidate actions. Because the execution layer is built on MoveIt, porting to other manipulators is straightforward; extending the supported task repertoire, by contrast, remains an open engineering effort that we leave to future work. 

UNCOM analyzes the recorded audio and video files, understands them, and sends the $x, y$ coordinates of the object and target in the image captured by the robot's camera, as well as the text of the requested task. Then, using the camera intrinsics and the depth image, the interaction node calculates the $x$, $y$ and $z$ pose of the object and target with respect to the robot's base. It tries to match the detected bounding boxes with the object detection from the image modules of the iChores framework~\cite{ichores}, to have proper 6DoF pose and mesh estimation. Otherwise, the system tries to grasp objects at the previously estimated center point. 

While the evaluation setup assumes the robot starts at an ideal distance from the desk, it is capable of navigating the environment as needed, but only moving forward and backwards is implemented in the current execution sub-system. While it does not require any markers over the table to work, the system requires that the robot returns to its initial pose every time for detection and object pose estimation consistency.  


The robot was able to understand instructions for the six following  scenarios: i) place a banana inside a bowl (no gesture guidance), ii) place a banana inside a bowl (no gesture guidance, one confounding object), iii) pick and place a lemon inside a bowl (with gesture and other objects), iv) place a lemon inside a bowl (multiple yellow fruits, 2 bowls, with gestures); v) place a banana on the side of a box (3 bananas, 1 bowl and a position relative to an object as target); and, finally, vi) object referred through deixis, target as an empty area.  

\begin{figure}
    \centering
    \vspace{.25 cm}
    \includegraphics[width=0.475\linewidth]{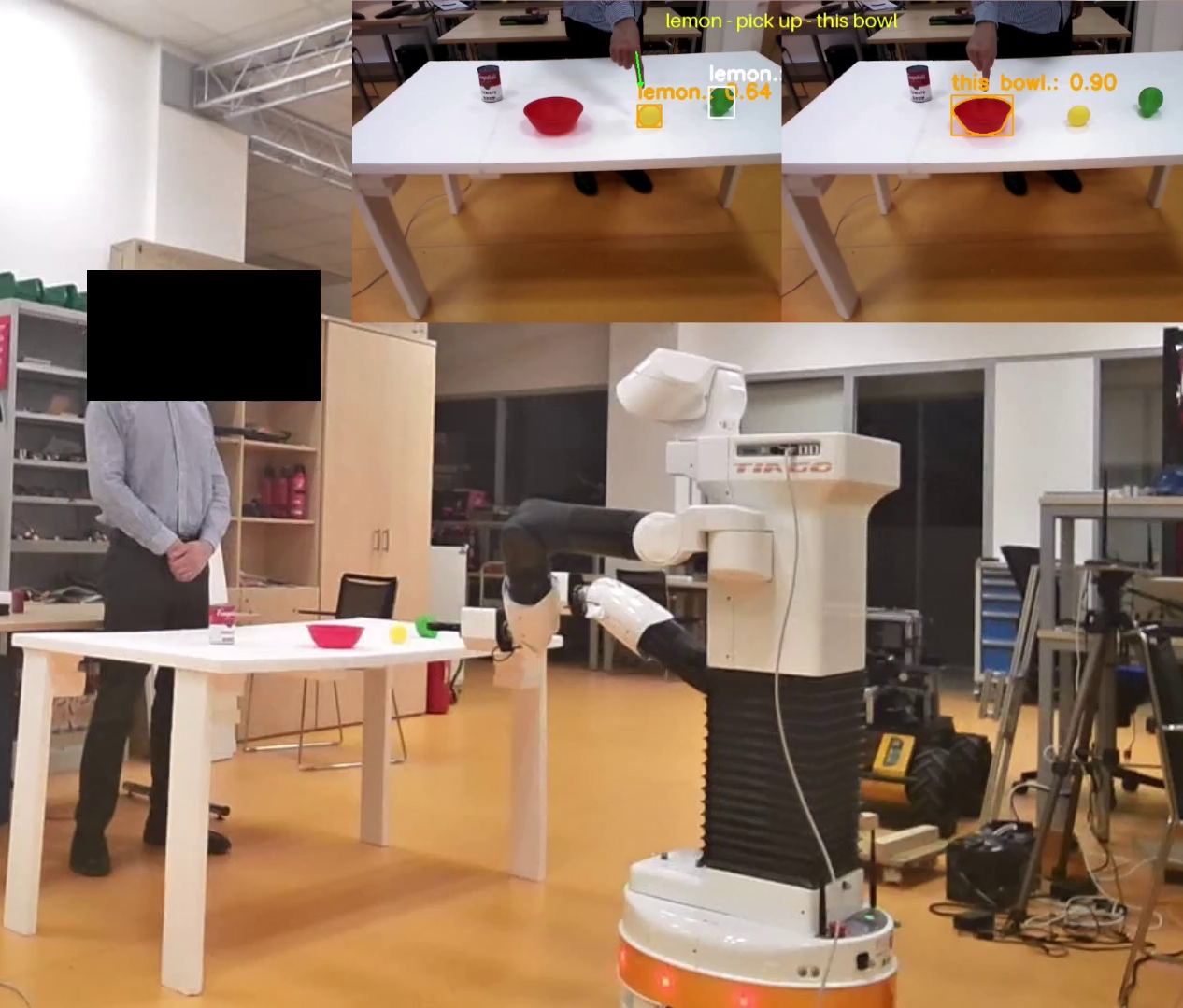}~\includegraphics[width=0.475\linewidth]{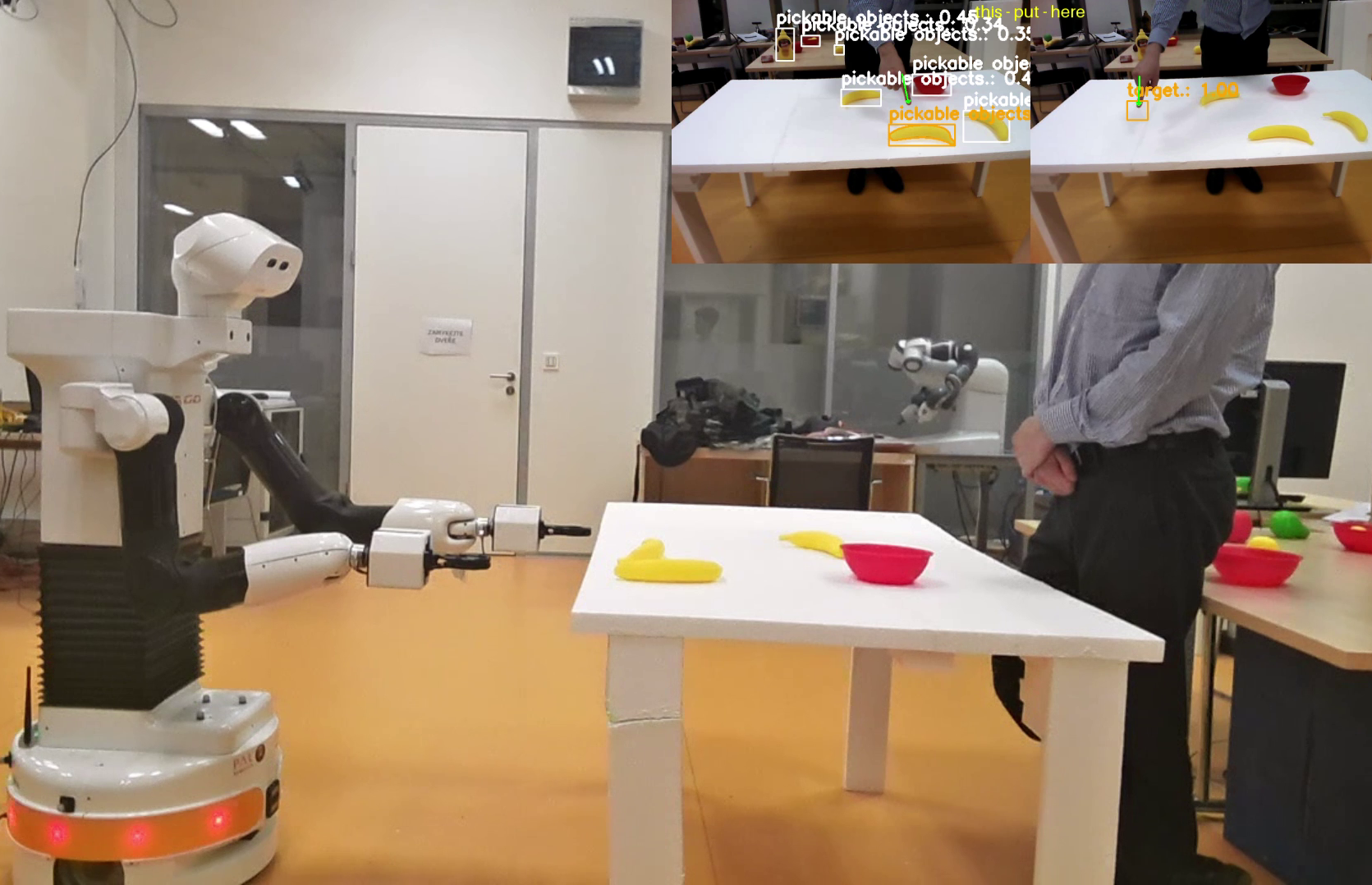}
    \caption{Outcomes of command comprehension and object/target-pointing behavior used for confirmation in scenarios III (top) and VI (bottom).}
    \label{fig:robot-demo}
    \vspace{-.5cm}
\end{figure}

Videos of the aforementioned demonstrations can be seen in \href{https://github.com/ichores-research/uncom/tree/ichores_pipeline/demo_videos}{this link}, and the results for scenarios iii) and iv) are shown in the top and bottom images of Figure~\ref{fig:robot-demo}.

\subsection{Algorithm performance evaluation}\label{subsec:benchmarking}

Even in constrained settings such as tabletop scenarios, human communication varies -- people use different words, gestures, and spatial references. Our algorithm handles this by combining speech and gestures, leveraging zero-shot deep learning models  to interpret commands without prior training. Table \ref{variations} summarizes the scenarios handled. The complete test data set is available online \footnote{https://huggingface.co/datasets/lubiluk/uncom}.

Users may refer to objects using nouns, adjectives, or deictic cues such as pointing, and words such as `this' or `that'. Scenes may include distractors, objects visually similar to the target, or unrelated clutter. Targets can be named objects or designated areas, identified absolutely (e.g. by pointing) or relatively (e.g. `next to the bottle'). Like objects, targets may have distractors, such as other regions in the same class.

The evaluation involved testing the algorithm on real-world examples recorded with a mobile phone camera. The videos were segmented into individual mp4 commands, covering the full range of variation that the algorithm was built to handle. Table \ref{examples} summarizes representative scenarios, and Figure \ref{fig:examples} shows annotated frames from the recordings.

Performance was measured by comparing the predicted actions with the intention of ground truth. A trial was considered successful when our system correctly identified the object, action, and target intended by the human instructor. Multiple takes were used for some commands to minimize the influence of errors from deep learning components (e.g., voice misrecognition), isolating the robustness of the procedural logic.

\begin{table}[h]
\vspace{-0.25cm}
\caption{Types of Variations Addressed by the Algorithm}
\label{variations}
\vspace{-0.25cm}
\begin{center}
\begin{tabular}{|c|c|}
\hline
Variation Subject & Possible Variants \\
\hline
\hline
Object & Reference, Deixis \\
Object distractors & Yes, No \\
Target & Reference, Deixis, Absolute Area, Relative Area \\
Target distractors & Yes, No \\
Additional clutter & Yes, No \\
\hline
\end{tabular}
\vspace{-.75cm}
\end{center}
\end{table}

\begin{table*}[h]
\vspace{.5cm}
\caption{Representative Example Scenarios for Evaluation}
\vspace{-.5cm}
\label{examples}
\begin{center}
\begin{tabular}{|c|c|c|c|c|c|}
\hline
Command & Object Type & \makecell{Object\\ Distractors} & Target Type & \makecell{Target\\ Distractors} & Clutter \\
\hline
\hline
Take the mug [pointing] and put it on this plate [pointing]  & Reference & No & Reference & No & No \\
Take the mug [pointing] and put it on this plate [pointing]  & Reference & Yes & Reference & No & No \\
Take this mug [pointing] and put it on this plate [pointing]  & Reference & Yes & Reference & Yes & No \\
Take the mug [pointing] and put it on this plate [pointing]  & Reference & No & Reference & No & No \\
Take this mug [pointing ] and put it here [pointing]  & Reference & No & Absolute Area & No  & No\\
Take this thing [pointing] and put it on this thing [pointing]  & Deixis & Yes & Deixis & Yes  & No\\
Take the mug [pointing ] and put it next to the plate [pointing]  & Reference & No & Relative Area & No & No \\
Take this plate [pointing] and stack it on top of the other plate [pointing]  & Reference & Yes & Reference & Yes & No \\
Take the banana [pointing] and put it inside of the frying pan [pointing]  & Reference & Yes & Reference & No & Yes \\
Take this fruit [pointing] and put it inside of this thing [pointing]  & Deixis & Yes & Deixis & No & Yes \\
Pour the cereal [pointing] into the bowl [pointing]  & Reference & Yes & Reference & No & Yes \\
\hline
\end{tabular}
  \vspace{-.75cm}
\end{center}
\end{table*}

\section{Results}

The algorithm output aligned closely with the intentions of the human instructor in almost all cases, confirming that the designed procedural logic functioned as intended. In particular, the algorithm successfully worked for all 22 benchmarking scenarios, albeit not for all videos. Upon aggregation of all the examples, it achieved an overall success rate of $82.39\%$ (131 of 159 instances).

However, several detection errors stemmed from limitations inherent in the underlying deep learning components. The MediaPipe hand detector proved to be the most vulnerable, exhibiting high sensitivity to motion blur and frequently missing hand detections. Additionally, the Whisper ASR model occasionally produced wrong transcriptions, such as repeatedly confusing the word `bowl' with `ball'. It is important to consider that the non-native English proficiency of the experimenters may have contributed to these transcription errors. In contrast, the GroundingDINO, Segment Anything, and DINOv2 models showed robust performance, with infrequent errors that were often correctable through contextual cues such as gesture information.

In general, while the algorithm effectively integrates multiple deep learning models to improve accuracy, the current performance bottleneck lies predominantly in these models themselves. Future improvements in model robustness will be essential for deploying such hybrid approaches in practical robotic systems.

\section{Conclusions \& Future Work}

This work introduced a command understanding algorithm tailored for tabletop manipulation tasks, integrating natural language commands with pointing gestures. The algorithm demonstrated a strong accuracy in extracting the core elements of commands—objects, actions, and targets—when recognition errors were minimal or compensated by contextual information such as gestures. Its framework effectively used the strengths of multiple deep learning models to address various input scenarios, achieving an overall success rate that exceeds 80\%, requiring only one specialist model for pointing detection.

The evaluation highlighted key challenges, mainly related to the limitations of the underlying models. The MediaPipe Hand Landmarker was particularly affected by motion blur, impairing reliable hand detection, while the Whisper ASR occasionally generated transcription errors, especially when processing speech from non-native English speakers. Despite these issues, the GroundingDINO, Segment Anything, and DINOv2 models consistently contributed to accurate perception and segmentation tasks. This reinforces that the effectiveness of the hybrid approach is tightly coupled to the quality of the component models.

A major strength of the proposed system is its modularity, allowing straightforward updates or replacement of individual models as advancements arise. As speech recognition and hand detection models continue to improve, we anticipate significant gains in overall system reliability and applicability to real-world robotic platforms.

Future work will prioritize improving the robustness of the algorithm against recognition errors and extending its ability to process continuous unconstrained data streams. The integration of command representation into a complete robotic control pipeline will be explored to enable the execution of the interpreted commands in dynamic environments.

In addition, expanding language support beyond English is a promising direction. This would mainly involve adapting the ASR and language understanding models to handle multiple languages, while preserving the existing decision-making architecture.

It is also necessary to establish a disambiguation routine in which the system can hold a conversation with users in order to ask for further clarification, until it can generate a feasible and successful actionable command for the robot.

\section*{Acknowledgments}

This research was supported in part by the National Science Center, Poland, under the OPUS call in the Weave program under the project number 2021/43/I/ST6/02489, and
from the funds assigned
to AGH University by the Polish Ministry of Education and Science.

We thank Prof. Michal Vavrečka of the Czech Institute of Informatics, Robotics and Cybernetics, CTU, Czech Republic, and Tessa Puli of the Automation and Control Institute (ACIN), TU Wien, Austria, for their valuable help in reviewing this work. We also thank Ahmad Ibrahim, Augustin Avalle, and Paul Wagner of the Institute of Computer Science of the Jagiellonian University, Poland, for their help in improving the pick-and-place pipeline used in this work. 

\bibliographystyle{IEEEtran}
\bibliography{IEEEabrv,bibliography}

\end{document}